\documentclass{article}
\usepackage{ijcai26}
\newcommand{\algname}{\textsc{SmoothVLA}}
\usepackage{times}
\usepackage{soul}
\usepackage{url}
\usepackage[hidelinks]{hyperref}
\usepackage[utf8]{inputenc}
\usepackage[small]{caption}
\usepackage{graphicx}
\usepackage{amsmath}
\usepackage{amsthm}
\usepackage{booktabs}
\usepackage{algorithm}
\usepackage{algorithmic}
\usepackage[switch]{lineno}

\usepackage{amssymb}
\usepackage{xcolor}
\usepackage{multirow}

\title{SmoothVLA: Aligning Vision-Language-Action Models with Physical Constraints via Intrinsic Smoothness Optimization}

\author{
Jiashun Li$^{1,2}$
\and
Xiaoyu Shi$^2$\and
Hong Xie$^3$\And
Mingsheng Shang$2$\And
Yun Lu$^2$\\
\affiliations
$^1$Chongqing University of Posts and Telecommunications;\\
$^2$Chongqing Institute of Green and Intelligent Technology, Chinese Academy of Sciences\\
$^3$The First Affiliated Hospital, University of Science and Technology of China\\
\emails
{}
}

\begin{document}

\maketitle

\begin{abstract}
Vision-Language-Action (VLA) models have emerged as a powerful paradigm for robotic manipulation. However, existing post-training methods face a dilemma between stability and exploration: Supervised Fine-Tuning (SFT) is constrained by demonstration quality and lacks generalization, whereas Reinforcement Learning (RL) improves exploration but often induces erratic, jittery trajectories that violate physical constraints. To bridge this gap, we propose SmoothVLA, a novel reinforcement learning fine-tuning framework that synergistically optimizes task performance and motion smoothness. The technical core is a physics-informed hybrid reward function that integrates binary sparse task rewards with a continuous dense term derived from trajectory jerk. Crucially, this reward is intrinsic, that computing directly from policy rollouts, without requiring extrinsic environment feedback or laborious reward engineering. Leveraging the Group Relative Policy Optimization (GRPO), SmoothVLA establishes trajectory smoothness as an explicit optimization prior, guiding the model toward physically feasible and stable control. Extensive experiments on the LIBERO benchmark demonstrate that SmoothVLA outperforms standard RL by 13.8\% in smoothness and significantly surpasses SFT in generalization across diverse tasks. Our work offers a scalable approach to aligning VLA models with physical-world constraints through intrinsic reward optimization. 

\end{abstract}

 \begin{figure}[!t]
    \centering
    \includegraphics[width=0.5\textwidth]{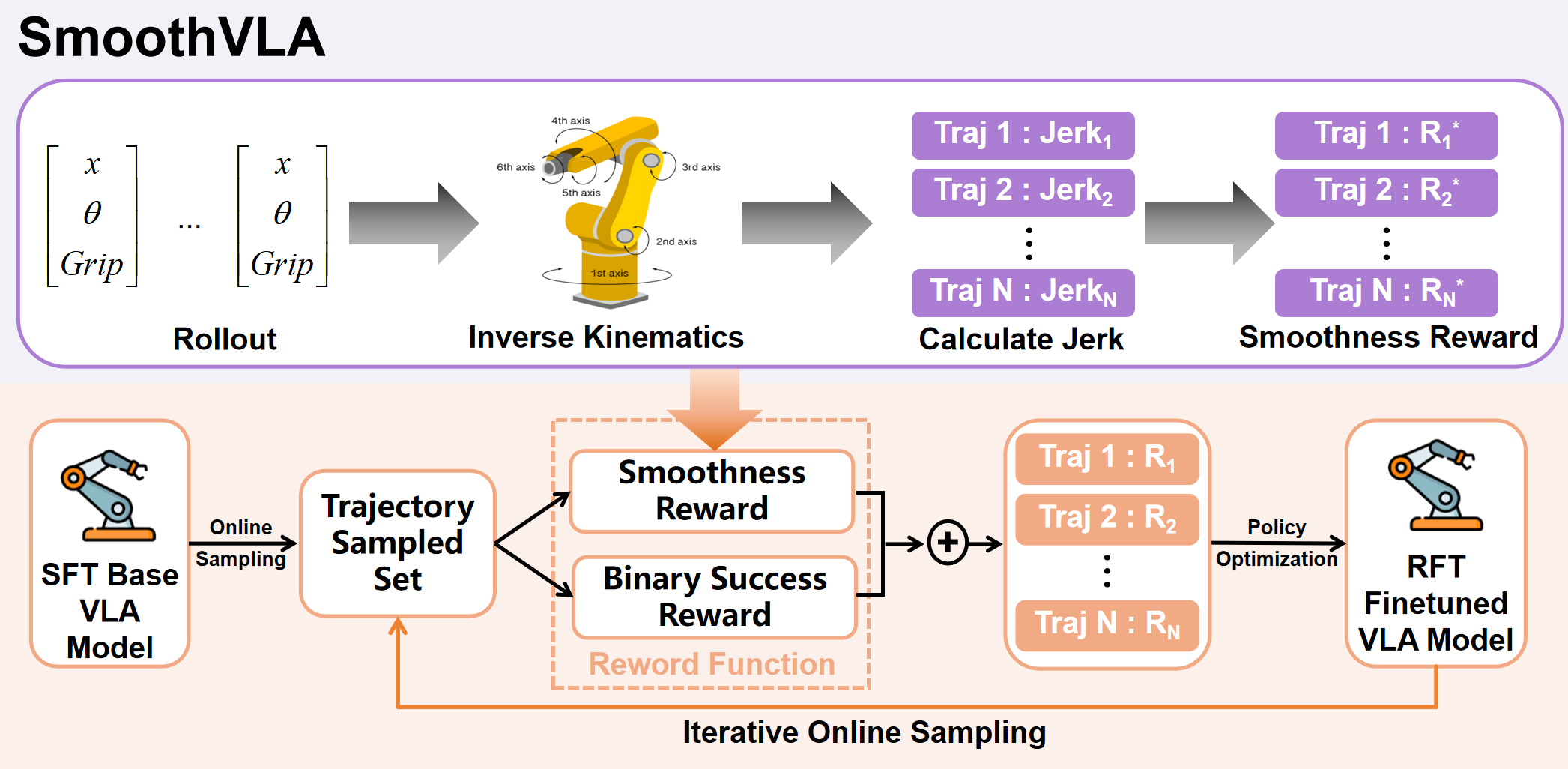}
    \caption{Overview of SmoothVLA.}
    \label{overview}
\end{figure}

\section{Introduction}

The rapid evolution of Vision-Language-Action (VLA) models has established a transformative paradigm for robotic manipulation. These models enable agents to interpret complex multimodal instructions and execute diverse tasks\cite{black2024pi_0,kim2024openvla,zitkovich2023rt}. However, a critical gap emerges as these models transition from simulations to open-ended physical environments. There is a clear misalignment between the high-dimensional outputs of neural networks and the fundamental kinematic constraints of physical hardware. In the physical world, trajectory smoothness is more than an aesthetic preference. It is a functional prerequisite\cite{12075}. Smooth trajectories ensure precise control, minimize mechanical wear, and guarantee operational safety\cite{flash1985coordination}. They serve as the essential bridge between digital reasoning and reliable physical execution.

Despite their potential, current VLA fine-tuning methodologies struggle to maintain this physical alignment. The dominant paradigm, Supervised Fine-Tuning (SFT), is inherently limited by the "demonstration bottleneck." It relies heavily on expensive, high-quality human data \cite{kim2502fine}. Consequently, SFT often fails to generalize when encountering out-of-distribution environmental perturbations. Alternatively, online Reinforcement Learning (RL) allows agents to transcend demonstration data through autonomous exploration\cite{li2025simplevla}. However, our analysis reveals an inherent "Exploration-Stability Paradox." While RL-driven exploration enhances generalization, the stochastic nature of policy searching frequently triggers high-frequency oscillations. These jittery motions degrade success rates and violate smooth-motion priors. This creates a \textit{physical misalignment} where the model's learned behavior becomes hardware-incompatible.

To bridge this gap and align VLA models with physical-world constraints, several non-trivial questions must be addressed: \begin{itemize} \item \textbf{How can we formalize motion quality into a differentiable objective} that captures the essence of robotic stability without oversimplifying the physics? \item \textbf{How can the model acquire physical feedback autonomously} from high-dimensional rollouts, eliminating the need for extrinsic sensors or manual reward engineering? \item \textbf{How can we optimize for smoothness without suppressing exploration} to avoid collapsing into \textit{safe but sub-optimal} policies? \end{itemize}

To answer these questions, we propose \textbf{SmoothVLA}, a reinforcement learning fine-tuning framework that jointly optimizes task performance and kinematic smoothness (Figure~\ref{overview}). The core innovation of SmoothVLA is a \textbf{physics-informed hybrid reward function} that augments sparse task rewards with a continuous dense term derived from trajectory \textbf{jerk} (the time derivative of acceleration)~\cite{schot1978jerk}. This intrinsic reward is computed directly from policy rollouts, enabling the model to \textit{self-correct} motion quality during exploration.
Starting from an SFT-tuned VLA model, the agent collects trajectory rollouts through online interaction. These trajectories are mapped to joint space via inverse kinematics, from which jerk is computed to quantify smoothness. The smoothness signal is then combined with the binary success reward to form a hybrid trajectory-level objective for reinforcement learning optimization. Through iterative sampling and policy updates, SmoothVLA progressively refines motion quality, aligning high-level policy outputs with low-level physical feasibility.
Our contributions are threefold:
\begin{itemize}
    \item We formally define the \textit{Exploration-Stability Paradox} in VLA reinforcement learning. By introducing trajectory \textbf{jerk} as a high-order differential bridge, we provide a mathematically rigorous framework to quantify the gap between neural actions and physically feasible motions.
    \item We propose a general-purpose RL post-training protocol to align VLA models with physical-world priors. The core is an \textbf{intrinsic, physics-informed reward mechanism} that enables autonomous self-correction. This framework is architecture-agnostic and offers a scalable solution for fine-tuning diverse embodied models without extrinsic environment modeling.
    \item We conduct rigorous evaluations on the \textbf{LIBERO} \cite{liu2023liberobenchmarkingknowledgetransfer} and \textbf{LIBERO-Plus} \cite{fei2025liberoplusindepthrobustnessanalysis} benchmarks. SmoothVLA establishes a new state-of-the-art in balancing task performance and execution quality. Our results show a \textbf{13.8\%} improvement in smoothness and significantly enhanced generalization, providing strong empirical evidence for physical-informed alignment.
\end{itemize}

\section{Empirical Study on Trajectory Smoothness in SFT and RL Fine-Tuned VLA Models}

To systematically evaluate how different fine-tuning paradigms affect the motion quality of trajectories generated by VLA models, and to quantify the core issue of “trajectory unsmoothness,” this chapter conducts a targeted empirical study. We consider a widely used robotic manipulation benchmark, LIBERO, and focus on a representative pick-and-place task: “pick up the alphabet soup and place it in the basket.” We evaluate the same base model (OpenVLA) fine-tuned with supervised fine-tuning and online reinforcement learning, and compare the generated trajectories from both qualitative visual inspection and quantitative metric analysis.

\subsection{Qualitative Analysis: Visual Comparison of Trajectory Smoothness}
\begin{figure}[htbp]
    \centering
    \includegraphics[width=0.5\textwidth]{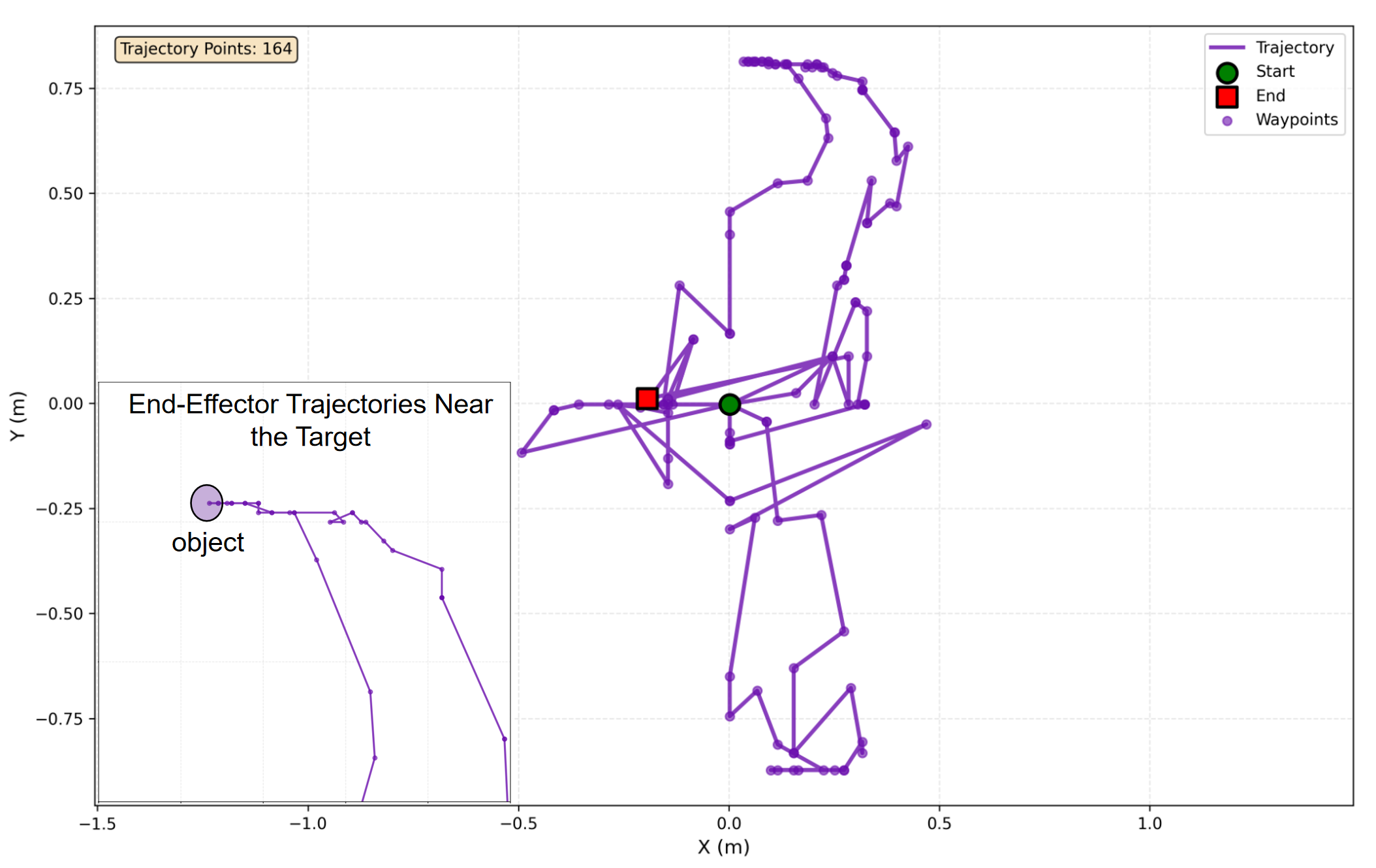}
    \caption{OpenVLA-SFT model}
    \label{sft}
\end{figure}

Figure\ref{sft} corresponds to the trajectory produced by the SFT OpenVLA model. As observed, the trajectory is globally coherent, with smooth curvature variations, and exhibits no abrupt direction changes or high-frequency oscillations during critical phases such as approaching, grasping, and transporting the object. Overall, it reflects intuitive and physically reasonable smooth motion.

\begin{figure}[htbp]
    \centering
    \includegraphics[width=0.5\textwidth]{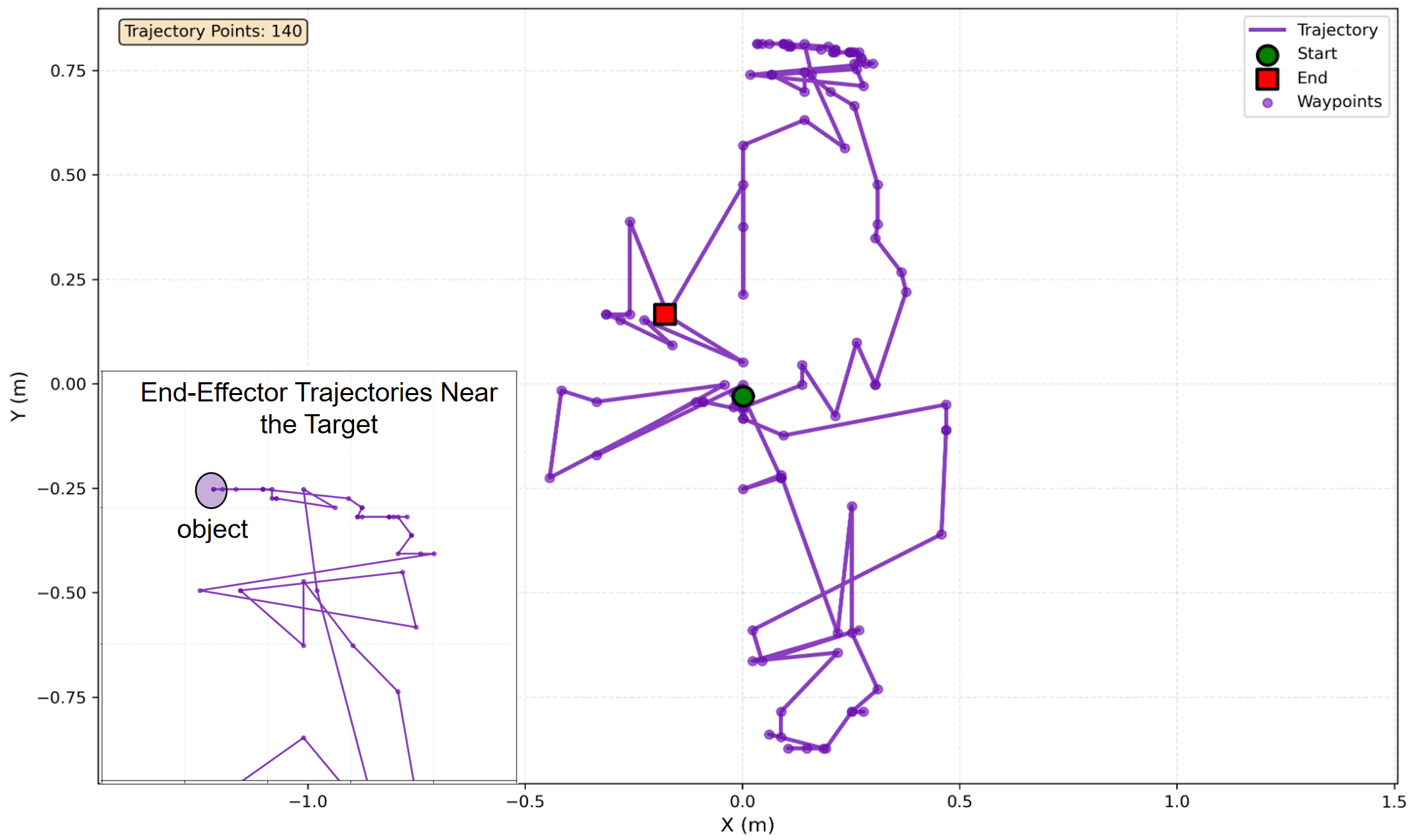}
    \caption{OpenVLA-RL model}
    \label{rl}
\end{figure}

In contrast, the trajectory generated by the Reinforcement-fine-tuned model in Figure\ref{rl} demonstrates pronounced unsmooth characteristics. The overall path contains multiple unnecessary detours and backtracking behaviors. More importantly, in the zoomed-in view near the target object (bottom-left inset), the trajectory exhibits severe local oscillations and jittering around the grasping point. Such discontinuous and unstable motion patterns not only increase the risk of grasp failure, but also indicate higher energy consumption and degraded dynamic stability.

\subsection{Quantitative Analysis: Jerk-based Metric}
To quantitatively validate these observations, we adopt jerk as the primary metric for evaluating trajectory smoothness. Jerk is defined as the time derivative of acceleration and physically reflects the rate of change of applied forces. In physical space, the smoothness of a trajectory is typically determined by the continuity of its kinematic characteristics. Jerk, defined as the third-order derivative of position with respect to time, directly reflects the rate of change of the applied force. In robotic motion, lower average jerk corresponds to smoother velocity profiles and more favorable dynamic properties\cite{mellinger2011minimum}. We conduct large-scale experiments on the LIBERO task suite and compute the average jerk of end-effector trajectories, with the results reported in Table \ref{tab:jerk_comparison}.

\begin{table}[htbp]
\centering
\small
\caption{Average Jerk Comparison In LIBERO}
\label{tab:jerk_comparison}
\begin{tabular}{c|ccccc}
\hline
\textbf{Model} & \textbf{Spatial} & \textbf{Object} & \textbf{Goal} & \textbf{Long} & \textbf{Average} \\
\hline
OpenVLA-SFT & 0.337 & 0.374 & 0.383 & 0.402 & 0.374 \\
OpenVLA-RL & 0.360 & 0.390 & 0.407 & 0.448 & 0.402 \\
\hline
\end{tabular}
\end{table}

The quantitative results are highly consistent with the qualitative observations. As shown in Table\ref{tab:jerk_comparison}, the Reinforcement-fine-tuned model produces trajectories with a significantly higher average jerk (0.402m/s³) than the Supervised-fine-tuned model (0.374m/s³). This provides strong physical evidence that existing online RL fine-tuning methods based on sparse rewards, while improving generalization, indeed introduce additional trajectory jitter and degrade motion smoothness.

\subsection{The Functional Correlation Between Trajectory Smoothness and Task Success Rate}

In the physical world, trajectory smoothness is not merely an aesthetic preference but a critical functional factor that directly affects task success. According to the minimum-jerk theory\cite{flash1985coordination}, smooth trajectories minimize abrupt variations in control inputs. In high-speed or high-precision tasks, non-smooth motions can induce instantaneous high jerk, which may demand torques beyond the physical bandwidth of the actuators, thereby causing system oscillations or severe tracking delays. Such effects are major contributors to task failures, including collisions and loss of stability.

Moreover, trajectory smoothness directly influences the accuracy of state estimation. Prior work shows that constraining jerk can effectively suppress residual vibrations at the end-effector, which is crucial for tasks involving precision assembly and object manipulation\cite{mellinger2011minimum}. When trajectories lack smoothness, violent motions can introduce substantial noise into sensors such as IMUs and vision systems, undermining the stability of closed-loop control and significantly degrading robustness and success rates in dynamic environments. Therefore, optimizing jerk is not only important for extending hardware lifespan, but also a necessary condition for ensuring that control commands translate into the intended physical behavior and for improving overall task success.

Taken together, these findings reveal a critical trade-off in VLA model fine-tuning. Supervised fine-tuning can learn smooth motion patterns from high-quality demonstrations, but its generalization ability is inherently constrained by the demonstration data distribution. Online reinforcement learning, by contrast, can enhance generalization through interaction-driven exploration; however, its intrinsic stochasticity and sparse reward structure often produce highly unsmooth trajectories, which in turn undermine control stability and directly reduce task success rates.The presence of this “smoothness–generalization” tension highlights the necessity and urgency of explicitly incorporating trajectory smoothness into the RL fine-tuning objective. This insight provides the fundamental motivation and empirical foundation for the proposed SmoothVLA framework.

\section{SmoothVLA}
\subsection{Preliminaries and Problem Formulation}
We consider the robotic manipulation task as a Markov Decision Process (MDP) defined by $(\mathcal{S}, \mathcal{A}, \mathcal{P}, \mathcal{R}, \gamma)$.
\subsubsection{State Space ($\mathcal{S}$):}The state $s_t \in \mathcal{S}$ is a multimodal tuple $s_{t} = (o_{t}^{\text{vis}}, o_{t}^{\text{prop}}, l_{\text{task}})$, integrating visual observations $o_{t}^{\mathrm{vis}}$ (e.g., RGB-D), proprioceptive states $o_{t}^{\mathrm{prop}}$ (e.g., joint positions $q_t$), and language instructions $l_{task}$.
\subsubsection{Action Space ($\mathcal{A}$):}The VLA policy $\pi_{\theta}$ maps $s_t$ to an action $a_t \in \mathbb{R}^d$. For typical end-effector control, $a_t$ represents the target pose or velocity.
\subsubsection{Action Chunking:}To ensure temporal coherence, the model generates a sequence (chunk) of actions $\mathbf{a}_{t:t+k} = \{a_t, \dots, a_{t+k-1}\}$. 

The trajectory $\tau$ is formed by the sequential execution of these actions under the system dynamics $\mathcal{P}$.

\subsection{Group Relative Policy Optimization (GRPO)}
To align VLA models with physical constraints without the computational burden of a critic network, we employ Group Relative Policy Optimization (GRPO). Unlike standard actor-critic RL, GRPO estimates the advantage function by comparing a group of $G$ trajectories $\{\tau_1, \dots, \tau_G\}$ sampled from the same instruction $l_{task}$.The optimization objective is:

\begin{multline}
\mathcal{J}(\theta) = \mathbb{E} \Bigg[ \frac{1}{G} \sum_{i=1}^G \Big( \min\big( \rho_i(\theta) \hat{A}_i, \\
\text{clip}\big(\rho_i(\theta), 1-\epsilon, 1+\epsilon\big) \hat{A}_i \big) - \beta D_{\text{KL}}(\pi_\theta \| \pi_{\text{ref}}) \Big) \Bigg]
\end{multline}

where $\rho_i(\theta)$ is the importance sampling ratio. The advantage $\hat{A}_i$ is computed by normalizing the total reward $R_i$ within the group: $\hat{A}_i = (R_i - \text{mean}(\mathbf{R})) / \text{std}(\mathbf{R})$. This intra-group competition is crucial for SmoothVLA, as it encourages the model to prefer the "smoothest" successful trajectory among multiple attempts.

\subsection{Physics-Informed Reward: From Jerk to Alignment}
The core of SmoothVLA is the transition from heuristic reward shaping to physical-informed alignment. We define a hybrid reward function that penalizes kinematic instability.The derivation of jerk is based on robotic differential kinematics theory. We first establish the fundamental velocity-level kinematic relationship:
\begin{equation}
\mathbf{V} = \mathbf{J}(\mathbf{q}) \dot{\mathbf{q}}
\end{equation}
where $\mathbf{V}$ is the end-effector velocity screw, $\mathbf{J}(\mathbf{q})$ is the Jacobian matrix that maps joint space to operational space, and $\dot{\mathbf{q}}$ is the joint velocity vector.

By differentiating the velocity relationship with respect to time, we obtain the complete acceleration expression:
\begin{equation}
\mathbf{A} = \dot{\mathbf{J}}(\mathbf{q}, \dot{\mathbf{q}}) \dot{\mathbf{q}} + \mathbf{J}(\mathbf{q}) \ddot{\mathbf{q}}
\end{equation}
where $\mathbf{A}$ is the end-effector acceleration screw, $\dot{\mathbf{J}}$ is the time derivative of the Jacobian matrix, and $\ddot{\mathbf{q}}$ is the joint acceleration vector.

Further differentiation of the acceleration expression yields the complete jerk relationship:
\begin{equation}
\mathbf{Jerk} = \ddot{\mathbf{J}}(\mathbf{q}, \dot{\mathbf{q}}, \ddot{\mathbf{q}}) \dot{\mathbf{q}} + 2\dot{\mathbf{J}}(\mathbf{q}, \dot{\mathbf{q}}) \ddot{\mathbf{q}} + \mathbf{J}(\mathbf{q}) \dddot{\mathbf{q}}
\end{equation}
where $\ddot{\mathbf{J}}$ is the second-order derivative of the Jacobian matrix, and $\dddot{\mathbf{q}}$ is the joint jerk vector.

Minimizing this term directly reduces mechanical vibration and ensures energy-efficient, human-like motion.
\subsubsection{The SmoothVLA Reward Function:}Inspired by recent advancements in outcome-driven alignment (e.g., DeepSeek-R1), we utilize a binary success indicator $I_{success}$ as the primary sparse signal. To inject the physical prior, we augment it with an intrinsic smoothness penalty:\begin{equation}R(\tau) = I_{success} \cdot \left( 1 - \lambda \cdot \frac{1}{T} \sum_{t=1}^{T} |\mathbf{Jerk}(t)|_{2} \right)\end{equation}where $\lambda$ is a scaling factor, $T$ is the total number of time steps in the trajectory. This design ensures that the agent first prioritizes task completion, and once successful, the GRPO objective drives the policy toward the most physically stable trajectory manifold. By computing Jerk directly from the policy's action sequence and proprioceptive feedback, SmoothVLA achieves feedback autonomy without extrinsic reward engineering.

\section{Experiment}
\subsection{Experimental Setups}
\subsubsection{Implementation Details} We employ OpenVLA as the backbone model, utilizing LoRA fine-tuning with the AdamW optimizer for both supervised and preference learning stages. During the supervised fine-tuning phase, we use a learning rate of $4 \times 10^{-5}$ with a batch size of 16. For the preference fine-tuning phase, we apply a learning rate of $2 \times 10^{-5}$ while maintaining the same batch size.

Specifically, in the reward function design, we introduce a smoothness weight coefficient $\lambda$ to control the relative importance of trajectory smoothness in the total reward. Through grid search and experimental validation, we ultimately set $\lambda =0.2$, which effectively balances the trade-off between task success rate and motion smoothness. This parameter configuration ensures that the policy maintains high task completion rates while generating physically feasible smooth motion trajectories.

Detailed parameter configurations for the training process and data preprocessing methods are provided in Appendices A and B. All experiments were conducted under identical hardware environments to ensure result comparability and reproducibility.

\subsection{Main Results on Robotics Benchmarks}
\subsubsection{Benchmarks} 
We evaluate SmoothVLA on the widely used simulation benchmark LIBERO. LIBERO is a lifelong learning benchmark focused on language-guided manipulation tasks across diverse object types, task specifications, and environments. It consists of five task suites: LIBERO-Goal, LIBERO-Spatial, LIBERO-Object, LIBERO-Long (10 tasks, each with 50 expert demonstrations).

\begin{figure}[htbp]
    \centering
    \includegraphics[width=0.5\textwidth]{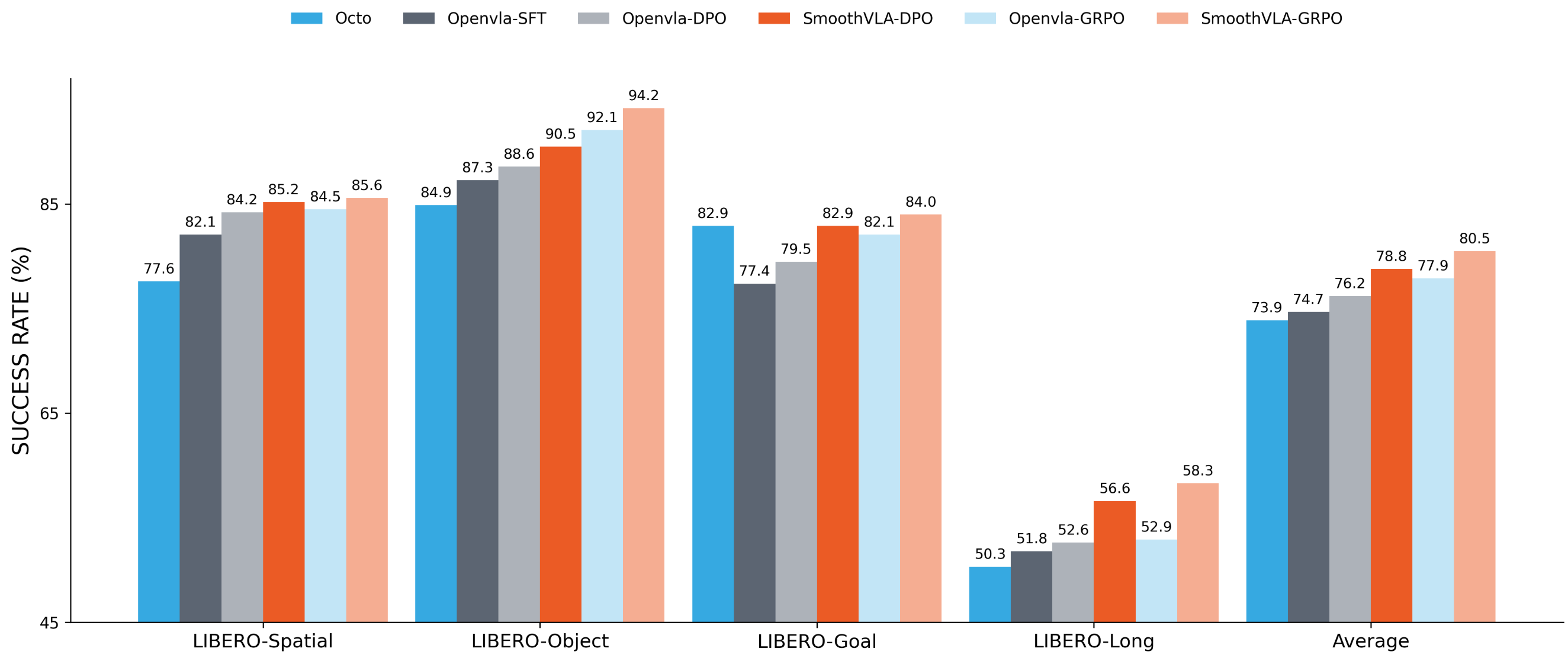}
    \caption{Comparison of SmoothVLA with OpenVLA and Octo fine-tuned on the same data on the LIBERO environment.}
    \label{fig1}
\end{figure}

\begin{figure}[htbp]
    \centering
    \includegraphics[width=0.5\textwidth]{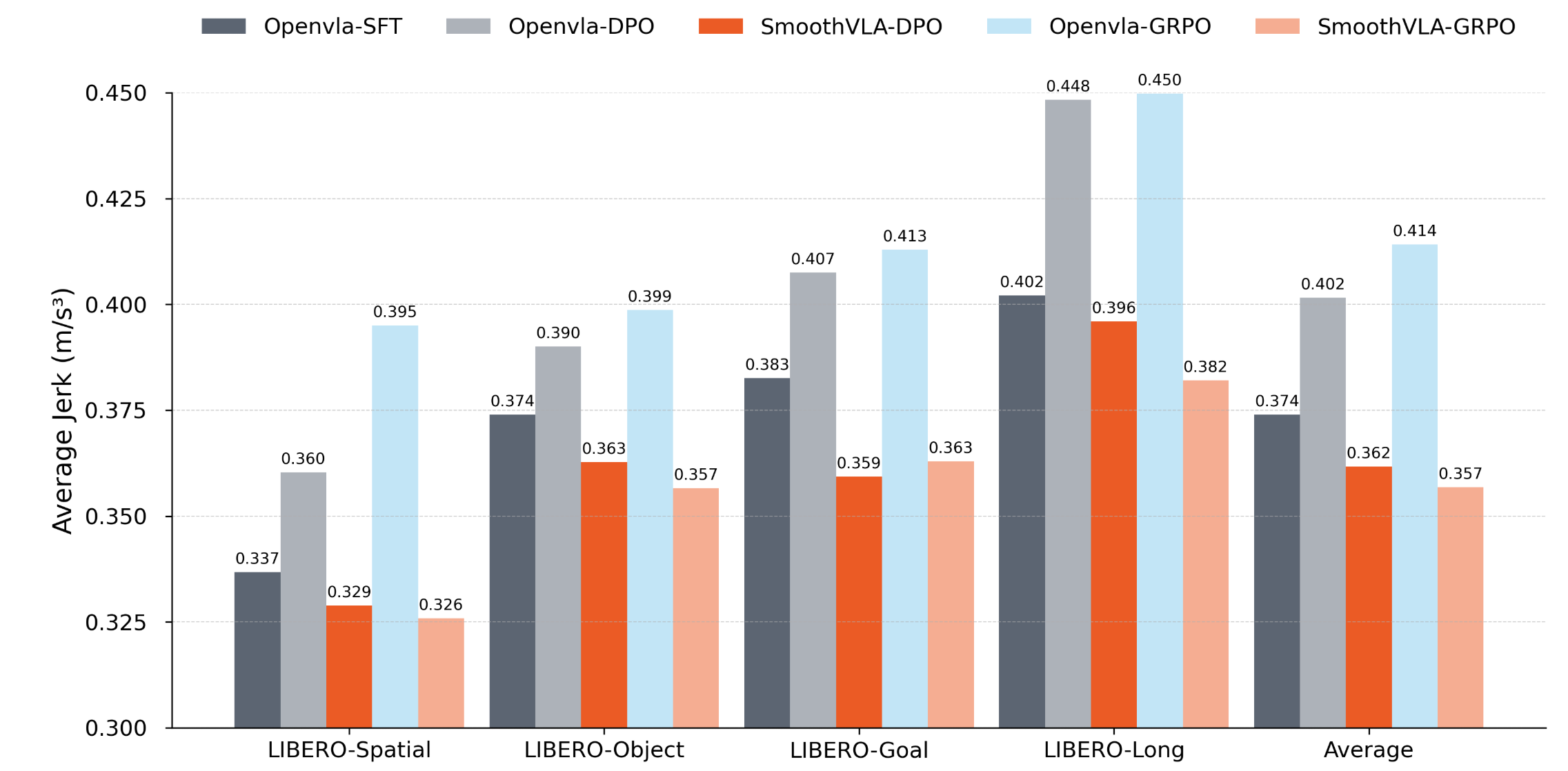}
    \caption{Comparison of motion smoothness through average jerk analysis between OpenVLA and SmoothVLA fine-tuned models evaluated on the LIBERO benchmark.}
    \label{fig2}
\end{figure}

\subsubsection{Robustness}
To comprehensively evaluate the robustness of SmoothVLA under realistic environmental variations, we employ the LIBERO-Plus benchmark platform for systematic validation. LIBERO-Plus is an extended version of the original LIBERO environment, specifically designed for in-depth analysis of model generalization capabilities under diverse perturbations. The results are shown in Table \ref{tab:results}.

In our experiments, we focus on evaluating four critical perturbation dimensions: Language Instructions (testing linguistic robustness through rewritten task descriptions), Light Conditions (varying illumination intensity, direction, and shadow patterns), Background Textures (modifying scene materials and appearances), and Objects Layout (adding distracting objects or altering target object positions). These tests are conducted on the same four task suites as in the benchmark study—LIBERO-Goal, LIBERO-Spatial, LIBERO-Object, and LIBERO-Long—ensuring consistency and enabling direct performance comparison.

\begin{table*}[htbp]
    \centering
    \caption{Comparison of model robustness across five perturbation types on the LIBERO-PLUS benchmark.}
    \label{tab:results}
    \begin{tabular}{lcccccc}
        \toprule
        Method & Original & Language & Light & Background & Layout & Average \\
        \midrule
        \multirow{2}{*}{OpenVLA-SFT} & 82.1 & 26.8 & 4.4 & 25.3 & 31.6 & 22.0 \\
         &  & $\downarrow$ 55.3 & $\downarrow$ 77.7 & $\downarrow$ 56.8 & $\downarrow$ 50.5 & $\downarrow$ 60.1 \\
        \midrule
        \multirow{2}{*}{OpenVLA-DPO} & 84.2 & 39.9 & 16.5 & 42.1 & 58.2 & 39.2 \\
         &  & $\downarrow$ 44.3 & $\downarrow$ 67.7 & $\downarrow$ 42.1 & $\downarrow$ 26.0 & $\downarrow$ 45.0 \\
        \midrule
        \multirow{2}{*}{SmoothVLA-DPO} & 85.2 & 40.1 & 17.5 & 58.6 & 62.9 & 44.8 \\
         &  & $\downarrow$ 45.1 & $\downarrow$ 67.7 & $\downarrow$ 26.6 & $\downarrow$ 22.3 & $\downarrow$ 40.4 \\
        \midrule
        \multirow{2}{*}{OpenVLA-GRPO} & 84.5 & 36.7 & 15.4 & 48.8 & 55.7 & 39.2 \\
         &  & $\downarrow$ 47.8 & $\downarrow$ 69.1 & $\downarrow$ 35.7 & $\downarrow$ 28.8 & $\downarrow$ 45.3 \\
        \midrule
        \multirow{2}{*}{SmoothVLA-GRPO} & 85.6 & 43.3 & 16.1 & 61.2 & 64.4 & 46.2 \\
         &  & $\downarrow$ 42.3 & $\downarrow$ 69.5 & $\downarrow$ 24.4 & $\downarrow$ 21.2 & $\downarrow$ 39.4 \\
        \bottomrule
    \end{tabular}
\end{table*}

\subsubsection{Results}
Comprehensive experimental results from Figure \ref{fig1}, Figure \ref{fig2}, and Table \ref{tab:results} demonstrate the significant advantages of our proposed SmoothVLA method across multiple dimensions. As shown in Figure \ref{fig1}, in the in-distribution benchmark evaluation, SmoothVLA-GRPO achieves an average success rate of 80.5\%, representing a 6.6 percentage point improvement over the baseline method Octo (73.9\%). This enhancement proves the effectiveness of SmoothVLA as a universal plugin—when applied to different reinforcement fine-tuning methods (DPO and GRPO) with the incorporation of trajectory smoothness evaluation, the average success rate is further increased by 2.6\%, highlighting its compatibility and performance gains in enhancing existing frameworks.

In the smoothness analysis of Figure \ref{fig2}, the SmoothVLA method significantly optimizes motion trajectory quality. Specifically, compared to traditional supervised fine-tuning methods, the smoothness metric improves by 4.5\%; when compared to standard RL fine-tuning methods, the improvement reaches 13.8\%. These results validate the effectiveness of the hybrid reward function in balancing task success and motion smoothness, demonstrating that SmoothVLA generates more stable and physically feasible trajectories.

The robustness evaluation in Table \ref{tab:results} further confirms the superiority of the method. SmoothVLA leads supervised fine-tuning methods by 24.2\% in average success rate, with variants based on GRPO and DPO leading by 7\% and 5.6\%, respectively. More importantly, when facing out-of-distribution perturbations, SmoothVLA exhibits smaller performance degradation, indicating stronger generalization capability and stability. These results collectively show that SmoothVLA not only enhances baseline performance but also improves model adaptability in complex environments through smoothness constraints.

\subsection{Ablation Study of Reward Model}
\begin{table}[H]
\centering
\caption{Ablation study of reward score}
\label{tab:performance_comparison}
\begin{tabular}{l*{5}{c}c}
\toprule
Model & \multicolumn{5}{c}{LIBERO-SPATIAL} & Avg \\
\cmidrule(lr){2-6}
 & In-dom & Lang & Light & Bkgd & Layout & \\
\midrule
$R_{\text{binary}}$ & 84.5 & 36.7 & 15.4 & 48.8 & 55.7 & 48.2 \\
$\Delta$ & -1.1 & -6.6 & -0.7 & -12.4 & -8.7 & -5.9 \\
\addlinespace
$R_{\text{random}}$ & 83.2 & 34.5 & 14.9 & 47.9 & 51.8 & 46.5 \\
$\Delta$ & -2.4 & -8.8 & -1.2 & -13.3 & -12.6 & -7.6 \\
\addlinespace
$R_{\text{smooth}}$ & 85.6 & 43.3 & 16.1 & 61.2 & 64.4 & 54.1 \\
\bottomrule
\end{tabular}
\end{table}
To systematically evaluate the impact of different reward function designs on model performance, we designed a rigorous ablation experimental protocol in Table \ref{tab:performance_comparison}. This study focuses on comparing the performance differences of three distinct reward function configurations in the LIBERO-Spatial environment, aiming to validate the critical role of trajectory smoothness reward terms in enhancing model generalization capability. The experiment adopts a controlled variable approach, maintaining identical training hyperparameters while only modifying the reward function design. The first configuration uses the basic binary sparse reward ($R_{\text{binary}}$), providing a reward of 1 only upon task success and 0 upon failure; the second configuration adds random noise terms to the binary reward ($R_{\text{random}}$), testing the impact of arbitrary reward perturbations on performance; the third configuration employs our proposed complete smoothness reward ($R_{\text{smooth}}$), which incorporates a jerk-based trajectory smoothness evaluation term on top of the binary reward. All experiments are conducted in the LIBERO-Spatial environment, which contains 10 complex tasks requiring spatial reasoning abilities and imposes high demands on motion trajectory precision and smoothness, effectively highlighting the performance differences among various reward function designs. The evaluation process covers five key dimensions: In-domain performance assesses basic capabilities under standard conditions; Language perturbation tests instruction understanding and execution consistency; Light variation examines visual perception robustness; Background interference evaluates scene adaptation capability; and Layout modification verifies spatial reasoning stability. Each experimental setup is independently run 5 times, reporting average performance metrics to ensure statistical significance of the results.

The results indicate that: (1) incorporating the smoothness reward term $R_{\text{smooth}}$ significantly enhances overall performance compared to basic binary rewards and random reward perturbations; (2) all reward components contribute substantially to model robustness across different perturbation types. These findings align with our theoretical expectations.
Specifically, the smoothness reward enhances the physical feasibility of generated trajectories by explicitly optimizing motion continuity. In parallel, the binary success reward $R_{\text{binary}}$ provides essential task-completion guidance, ensuring the model maintains high success rates in in-domain scenarios. The significant performance gap between $R_{\text{smooth}}$ and $R_{\text{random}}$ further demonstrates that structured physical priors are more effective than arbitrary reward perturbations.
Notably, the smoothness reward shows particularly strong improvements in layout variation tasks, indicating its importance in spatial reasoning adaptation. Similarly, the 6.6-point improvement in language perturbation tasks suggests that motion smoothness contributes to better instruction grounding and execution consistency.

\subsection{Case Study}

\begin{figure}[htbp]
    \centering
    \includegraphics[width=0.5\textwidth]{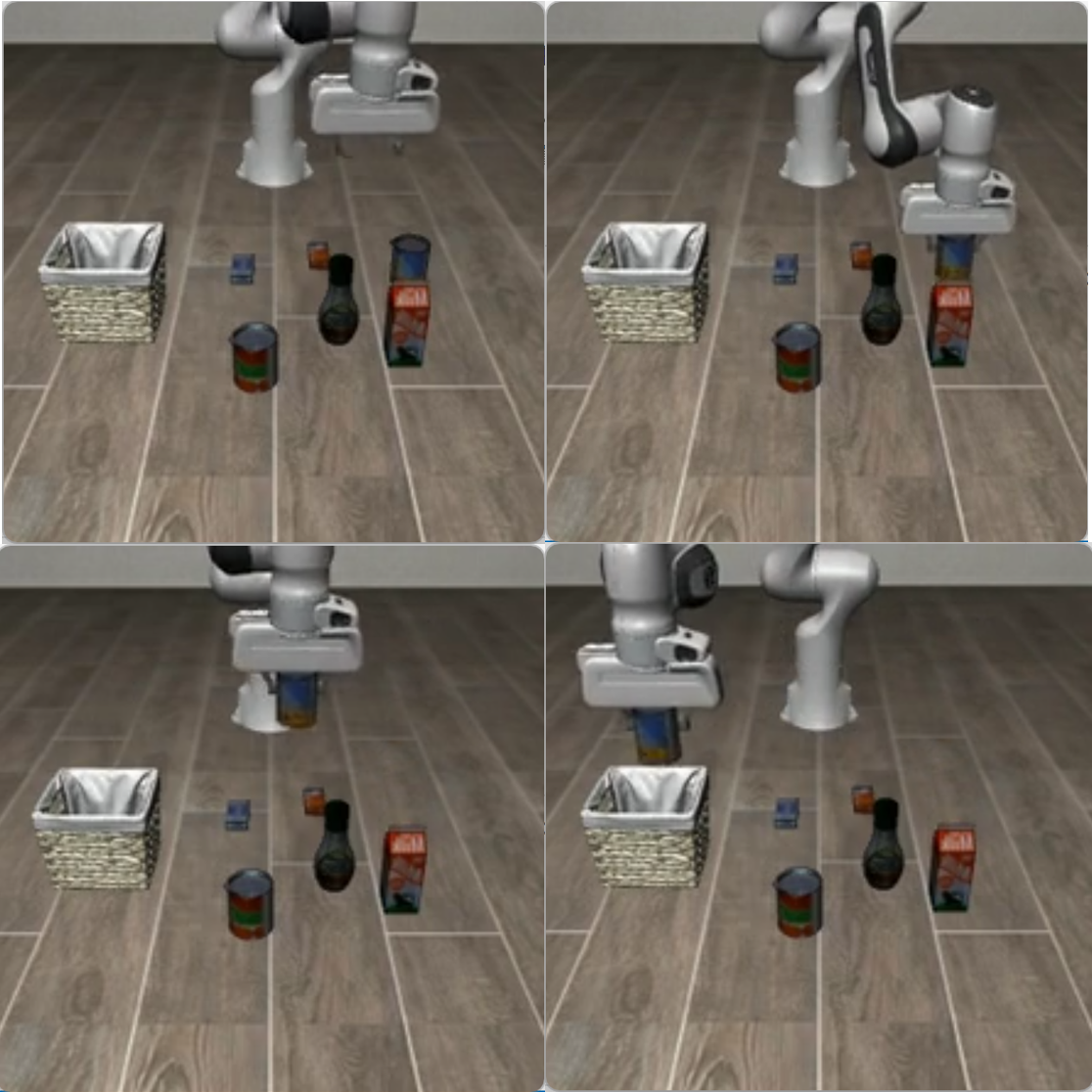}
    \caption{Scene diagram in LIBERO}
    \label{scene}
\end{figure}

\begin{figure}[htbp]
    \centering
    \includegraphics[width=0.5\textwidth]{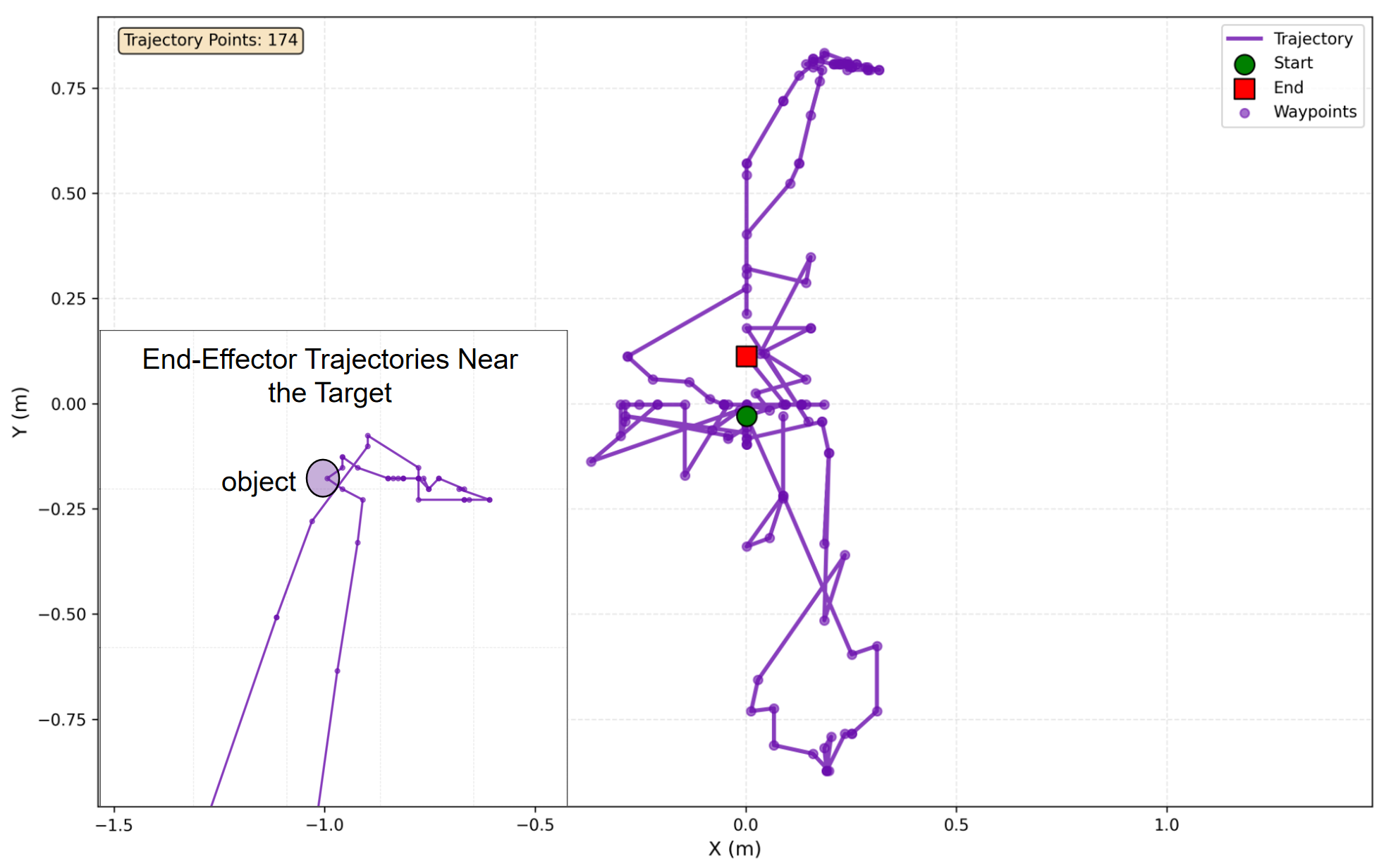}
    \caption{Trajectory case by SmoothVLA}
    \label{smooth}
\end{figure}

We further present a case study to analyze the advantages of SmoothVLA in terms of trajectory quality. The case is drawn from a representative task in the LIBERO benchmark, namely “pick up the alphabet soup and place it in the basket” (Figure \ref{scene}). This task requires the robot to accurately place the alphabet soup can into the basket while avoiding collisions with surrounding obstacles, thereby imposing stringent requirements on motion precision and trajectory smoothness.

The trajectory generated by the Supervised-fine-tuned model in Figure \ref{sft} is relatively smooth overall, but still exhibits certain limitations in complex interaction stages. In contrast, the trajectory produced by the RL-finetuned model in Figure \ref{rl} shows evident back-and-forth movements and high-frequency jitter, especially near the target object and during the placement phase, leading to unnecessary oscillations and degraded control stability.

Building upon this, the trajectory generated by SmoothVLA is significantly smoother than that of the RL baseline and exhibits almost no backtracking or jitter (Figure \ref{smooth}). During the placement process, the robotic arm maintains a stable posture and continuous motion, effectively avoiding collisions with surrounding obstacles and ensuring successful task completion, which demonstrates the superior motion control capability of SmoothVLA.

Further analysis suggests that the superiority of SmoothVLA stems from its smoothness-aware reward mechanism. By explicitly optimizing trajectory jerk, the model learns to generate physically more feasible motion trajectories. This optimization not only improves task success rates but also substantially enhances the robot’s motion performance. Such improvements are particularly evident in tasks requiring precise manipulation, validating the effectiveness of our method in improving the motion quality of VLA models.

\section{Related Works}

\subsubsection{Vision-Language-Action Models.}
Early robot learning systems commonly adopt hierarchical planning paradigms~\cite{huang2024rekep,li2024hamster,robocodex,huang2023voxposer,mu2024embodiedgpt}. In such frameworks, large language models or vision-language models are typically employed to produce high-level plans, while separate low-level controllers execute local motion trajectories. Representative examples include Code as Policies~\cite{liang2023code} and EmbodiedGPT~\cite{mu2024embodiedgpt}. Although effective in structured settings, these approaches often suffer from limited low-level motor competence and exhibit poor scalability to diverse real-world manipulation tasks.

In contrast, Vision-Language-Action (VLA) models aim to directly generate actions within an end-to-end architecture by scaling up low-level control with strong VLM backbones. Existing VLA systems generally fall into two main categories. 
The first category discretizes the continuous action space into symbolic tokens~\cite{kim2024openvla,brohan2023rt,brohan2022rt}. For instance, OpenVLA~\cite{kim2024openvla} maintains an autoregressive language modeling objective by representing actions as a limited vocabulary of \textit{action tokens}. However, such discretization inevitably introduces approximation errors, motivating subsequent works~\cite{black2024pi_0} to adopt alternative architectures~\cite{zhou2024transfusion} that incorporate diffusion-based heads for continuous action prediction.
The second category leverages diffusion models~\cite{chi2023diffusion,diffuser,adaptdiffuser,decisiondiffuser} as action generators. Methods such as Diffusion Policy~\cite{chi2023diffusion} produce future action sequences via iterative denoising, instead of predicting actions step-by-step in an autoregressive manner.

Despite architectural differences, most existing VLA models are trained using supervised learning on successful demonstration trajectories, typically through behavior cloning. This paradigm inherently limits their ability to generalize to unseen tasks and environments. In contrast, our \algname departs from purely supervised alignment and instead optimizes VLA policies directly at the trajectory level via trial-and-error interaction, substantially enhancing generalization and adaptability.

\subsubsection{Reinforcement Learning and Preference Optimization.}
Reinforcement learning (RL)~\cite{christiano2017deep,ziegler2019fine,schulman2017proximal} has become a cornerstone technique for post-training foundation models~\cite{dubey2024llama,achiam2023gpt,fan2024reinforcement,wang2025beyond}, particularly in aligning pre-trained models with human values through preference-based optimization.

RL has also demonstrated strong effectiveness in robotic policy learning~\cite{saferlchen,wang2024escirl,chen2021decision,RLdexterousReview,wu2024unidexfpm}. Nevertheless, successfully applying RL to fine-tune VLA models remains challenging. Prior work suggests several key obstacles: (1) manipulation tasks often involve complex, long-horizon objectives, making reward specification difficult~\cite{finn2016guided}; (2) while preference-based reward learning is feasible, collecting high-quality human preferences for robotic trajectories is typically expensive and time-consuming~\cite{walke2023bridgedata}; and (3) unstable or noisy reward gradients can easily cause policy optimization algorithms such as PPO~\cite{schulman2017proximal} to diverge~\cite{bucsoniu2018reinforcement}.

Recently, a line of work~\cite{rafailov2024direct,wang2024preference} has demonstrated that policies can be effectively aligned via RL-style optimization without explicit reward modeling, by directly contrasting trajectories. Inspired by these advances, \algname aligns VLA policies through trajectory-level comparison, thereby avoiding the need for handcrafted reward functions. Furthermore, we propose an automatic preference synthesis pipeline that naturally scales to diverse manipulation tasks and supports flexible alignment objectives.

\section{Conclusion}
In this paper, we investigated a fundamental yet underexplored issue in VLA model fine-tuning: the degradation of trajectory smoothness introduced by reinforcement learning, and its negative impact on motion quality and task performance. To address this challenge, we proposed SmoothVLA, a reinforcement learning fine-tuning framework that explicitly incorporates trajectory smoothness into the optimization objective. The core of our approach is a novel hybrid reward function that augments sparse task rewards with a dense jerk-based smoothness reward, which can be directly computed from rollout trajectories without additional supervision. Extensive experiments on multiple LIBERO benchmarks demonstrate that SmoothVLA consistently produces smoother and more physically feasible trajectories, while simultaneously improving task success rates and generalization performance. These results suggest that explicitly optimizing motion quality is a crucial component for reliable and robust RL fine-tuning of VLA models.

Despite the promising results, this work opens several directions for future research.First, we plan to extend SmoothVLA to more complex real-world robotic platforms and long-horizon manipulation tasks, where smoothness constraints are even more critical for safety and stability. Second, we aim to explore more adaptive and learnable smoothness objectives, such as task-dependent or perception-aware smoothness rewards, to further improve flexibility across diverse scenarios.Finally, we are interested in integrating trajectory smoothness optimization with model-based control or world-model learning, in order to bridge high-level policy learning and low-level motion feasibility more tightly.

\bibliographystyle{named}
\bibliography{ijcai26}

@article{black2024pi_0,
  title={$\pi_0$: A Vision-Language-Action Flow Model for General Robot Control},
  author={Black, Kevin and Brown, Noah and Driess, Danny and Esmail, Adnan and Equi, Michael and Finn, Chelsea and Fusai, Niccolo and Groom, Lachy and Hausman, Karol and Ichter, Brian and others},
  journal={arXiv preprint arXiv:2410.24164},
  year={2024}
}

@article{kim2024openvla,
  title={Openvla: An open-source vision-language-action model},
  author={Kim, Moo Jin and Pertsch, Karl and Karamcheti, Siddharth and Xiao, Ted and Balakrishna, Ashwin and Nair, Suraj and Rafailov, Rafael and Foster, Ethan and Lam, Grace and Sanketi, Pannag and others},
  journal={arXiv preprint arXiv:2406.09246},
  year={2024}
}

@inproceedings{zitkovich2023rt,
  title={Rt-2: Vision-language-action models transfer web knowledge to robotic control},
  author={Zitkovich, Brianna and Yu, Tianhe and Xu, Sichun and Xu, Peng and Xiao, Ted and Xia, Fei and Wu, Jialin and Wohlhart, Paul and Welker, Stefan and Wahid, Ayzaan and others},
  booktitle={Conference on Robot Learning},
  pages={2165--2183},
  year={2023},
  organization={PMLR}
}

@article{kim2502fine,
  title={Fine-tuning vision-language-action models: Optimizing speed and success, 2025},
  author={Kim, Moo Jin and Finn, Chelsea and Liang, Percy},
  journal={URL https://arxiv. org/abs/2502.19645}
}

@article{li2025simplevla,
  title={Simplevla-rl: Scaling vla training via reinforcement learning},
  author={Li, Haozhan and Zuo, Yuxin and Yu, Jiale and Zhang, Yuhao and Yang, Zhaohui and Zhang, Kaiyan and Zhu, Xuekai and Zhang, Yuchen and Chen, Tianxing and Cui, Ganqu and others},
  journal={arXiv preprint arXiv:2509.09674},
  year={2025}
}

@article{schot1978jerk,
  title={Jerk: the time rate of change of acceleration},
  author={Schot, Steven H},
  journal={American Journal of Physics},
  volume={46},
  number={11},
  pages={1090--1094},
  year={1978},
  publisher={American Association of Physics Teachers}
}

@article{brohan2023rt,
  title={Rt-2: Vision-language-action models transfer web knowledge to robotic control},
  author={Brohan, Anthony and Brown, Noah and Carbajal, Justice and Chebotar, Yevgen and Chen, Xi and Choromanski, Krzysztof and Ding, Tianli and Driess, Danny and Dubey, Avinava and Finn, Chelsea and others},
  journal={arXiv preprint arXiv:2307.15818},
  year={2023}
}

@inproceedings{adaptdiffuser,
  title={AdaptDiffuser: Diffusion Models as Adaptive Self-evolving Planners},
  author={Liang, Zhixuan and Mu, Yao and Ding, Mingyu and Ni, Fei and Tomizuka, Masayoshi and Luo, Ping},
  booktitle={International Conference on Machine Learning},
  pages={20725--20745},
  year={2023},
  organization={PMLR}
}

@inproceedings{diffuser,
  title={Planning with Diffusion for Flexible Behavior Synthesis},
  author={Janner, Michael and Du, Yilun and Tenenbaum, Joshua and Levine, Sergey},
  booktitle={International Conference on Machine Learning},
  pages={9902--9915},
  year={2022},
  organization={PMLR}
}

@inproceedings{robocodex,
  title={RoboCodeX: Multimodal Code Generation for Robotic Behavior Synthesis},
  author={Mu, Yao and Chen, Junting and Zhang, Qing-Long and Chen, Shoufa and Yu, Qiaojun and Chongjian, GE and Chen, Runjian and Liang, Zhixuan and Hu, Mengkang and Tao, Chaofan and others},
  booktitle={Forty-first International Conference on Machine Learning},
  year={2024}
}

@inproceedings{RLdexterousReview,
  title={Dexterous manipulation with deep reinforcement learning: Efficient, general, and low-cost},
  author={Zhu, Henry and Gupta, Abhishek and Rajeswaran, Aravind and Levine, Sergey and Kumar, Vikash},
  booktitle={2019 International Conference on Robotics and Automation (ICRA)},
  pages={3651--3657},
  year={2019},
  organization={IEEE}
}

@article{wu2024unidexfpm,
  title={Unidexfpm: Universal dexterous functional pre-grasp manipulation via diffusion policy},
  author={Wu, Tianhao and Gan, Yunchong and Wu, Mingdong and Cheng, Jingbo and Yang, Yaodong and Zhu, Yixin and Dong, Hao},
  journal={arXiv preprint arXiv:2403.12421},
  year={2024}
}

@article{huang2023voxposer,
  title={Voxposer: Composable 3d value maps for robotic manipulation with language models},
  author={Huang, Wenlong and Wang, Chen and Zhang, Ruohan and Li, Yunzhu and Wu, Jiajun and Fei-Fei, Li},
  journal={arXiv preprint arXiv:2307.05973},
  year={2023}
}

@inproceedings{decisiondiffuser,
  title={Is Conditional Generative Modeling all you need for Decision Making?},
  author={Ajay, Anurag and Du, Yilun and Gupta, Abhi and Tenenbaum, Joshua B and Jaakkola, Tommi S and Agrawal, Pulkit},
  booktitle={The Eleventh International Conference on Learning Representations},
  year={2023},
}

@inproceedings{liang2023code,
  title={Code as policies: Language model programs for embodied control},
  author={Liang, Jacky and Huang, Wenlong and Xia, Fei and Xu, Peng and Hausman, Karol and Ichter, Brian and Florence, Pete and Zeng, Andy},
  booktitle={2023 IEEE International Conference on Robotics and Automation (ICRA)},
  pages={9493--9500},
  year={2023},
  organization={IEEE}
}

@inproceedings{
wang2024escirl,
title={Esc{IRL}: Evolving Self-Contrastive {IRL} for Trajectory Prediction in Autonomous Driving},
author={Siyue Wang and Zhaorun Chen and Zhuokai Zhao and Chaoli Mao and Yiyang Zhou and Jiayu He and Albert Sibo Hu},
booktitle={8th Annual Conference on Robot Learning},
year={2024},
url={https://openreview.net/forum?id=1IzW0aniyg}
}

@article{schulman2017proximal,
  title={Proximal policy optimization algorithms},
  author={Schulman, John and Wolski, Filip and Dhariwal, Prafulla and Radford, Alec and Klimov, Oleg},
  journal={arXiv preprint arXiv:1707.06347},
  year={2017}
}

@article{chi2023diffusion,
  title={Diffusion policy: Visuomotor policy learning via action diffusion},
  author={Chi, Cheng and Xu, Zhenjia and Feng, Siyuan and Cousineau, Eric and Du, Yilun and Burchfiel, Benjamin and Tedrake, Russ and Song, Shuran},
  journal={The International Journal of Robotics Research},
  pages={02783649241273668},
  year={2023},
  publisher={SAGE Publications Sage UK: London, England}
}

@article{ziegler2019fine,
  title={Fine-tuning language models from human preferences},
  author={Ziegler, Daniel M and Stiennon, Nisan and Wu, Jeffrey and Brown, Tom B and Radford, Alec and Amodei, Dario and Christiano, Paul and Irving, Geoffrey},
  journal={arXiv preprint arXiv:1909.08593},
  year={2019}
}

@article{christiano2017deep,
  title={Deep reinforcement learning from human preferences},
  author={Christiano, Paul F and Leike, Jan and Brown, Tom and Martic, Miljan and Legg, Shane and Amodei, Dario},
  journal={Advances in neural information processing systems},
  volume={30},
  year={2017}
}

@article{mu2024embodiedgpt,
  title={Embodiedgpt: Vision-language pre-training via embodied chain of thought},
  author={Mu, Yao and Zhang, Qinglong and Hu, Mengkang and Wang, Wenhai and Ding, Mingyu and Jin, Jun and Wang, Bin and Dai, Jifeng and Qiao, Yu and Luo, Ping},
  journal={Advances in Neural Information Processing Systems},
  volume={36},
  year={2024}
}

@article{zhou2024transfusion,
  title={Transfusion: Predict the next token and diffuse images with one multi-modal model},
  author={Zhou, Chunting and Yu, Lili and Babu, Arun and Tirumala, Kushal and Yasunaga, Michihiro and Shamis, Leonid and Kahn, Jacob and Ma, Xuezhe and Zettlemoyer, Luke and Levy, Omer},
  journal={arXiv preprint arXiv:2408.11039},
  year={2024}
}

@inproceedings{walke2023bridgedata,
  title={Bridgedata v2: A dataset for robot learning at scale},
  author={Walke, Homer Rich and Black, Kevin and Zhao, Tony Z and Vuong, Quan and Zheng, Chongyi and Hansen-Estruch, Philippe and He, Andre Wang and Myers, Vivek and Kim, Moo Jin and Du, Max and others},
  booktitle={Conference on Robot Learning},
  pages={1723--1736},
  year={2023},
  organization={PMLR}
}

@article{brohan2022rt,
  title={Rt-1: Robotics transformer for real-world control at scale},
  author={Brohan, Anthony and Brown, Noah and Carbajal, Justice and Chebotar, Yevgen and Dabis, Joseph and Finn, Chelsea and Gopalakrishnan, Keerthana and Hausman, Karol and Herzog, Alex and Hsu, Jasmine and others},
  journal={arXiv preprint arXiv:2212.06817},
  year={2022}
}

@article{huang2024rekep,
  title={Rekep: Spatio-temporal reasoning of relational keypoint constraints for robotic manipulation},
  author={Huang, Wenlong and Wang, Chen and Li, Yunzhu and Zhang, Ruohan and Fei-Fei, Li},
  journal={arXiv preprint arXiv:2409.01652},
  year={2024}
}

@article{wang2025beyond,
  title={Beyond Reward Hacking: Causal Rewards for Large Language Model Alignment},
  author={Wang, Chaoqi and Zhao, Zhuokai and Jiang, Yibo and Chen, Zhaorun and Zhu, Chen and Chen, Yuxin and Liu, Jiayi and Zhang, Lizhu and Fan, Xiangjun and Ma, Hao and others},
  journal={arXiv preprint arXiv:2501.09620},
  year={2025}
}

@article{bucsoniu2018reinforcement,
  title={Reinforcement learning for control: Performance, stability, and deep approximators},
  author={Bu{\c{s}}oniu, Lucian and De Bruin, Tim and Toli{\'c}, Domagoj and Kober, Jens and Palunko, Ivana},
  journal={Annual Reviews in Control},
  volume={46},
  pages={8--28},
  year={2018},
  publisher={Elsevier}
}

@inproceedings{saferlchen,
      title={Safe Reinforcement Learning via Hierarchical Adaptive Chance-Constraint Safeguards}, 
      author={Zhaorun Chen and Zhuokai Zhao and Tairan He and Binhao Chen and Xuhao Zhao and Liang Gong and Chengliang Liu},
      year={2024},
      booktitle={IEEE/RSJ International Conference on Intelligent Robots and Systems (IROS)},
}

@inproceedings{finn2016guided,
  title={Guided cost learning: Deep inverse optimal control via policy optimization},
  author={Finn, Chelsea and Levine, Sergey and Abbeel, Pieter},
  booktitle={International conference on machine learning},
  pages={49--58},
  year={2016},
  organization={PMLR}
}

@article{rafailov2024direct,
  title={Direct preference optimization: Your language model is secretly a reward model},
  author={Rafailov, Rafael and Sharma, Archit and Mitchell, Eric and Manning, Christopher D and Ermon, Stefano and Finn, Chelsea},
  journal={Advances in Neural Information Processing Systems},
  volume={36},
  year={2024}
}

@article{fan2024reinforcement,
  title={Reinforcement learning for fine-tuning text-to-image diffusion models},
  author={Fan, Ying and Watkins, Olivia and Du, Yuqing and Liu, Hao and Ryu, Moonkyung and Boutilier, Craig and Abbeel, Pieter and Ghavamzadeh, Mohammad and Lee, Kangwook and Lee, Kimin},
  journal={Advances in Neural Information Processing Systems},
  volume={36},
  year={2024}
}

@article{dubey2024llama,
  title={The llama 3 herd of models},
  author={Dubey, Abhimanyu and Jauhri, Abhinav and Pandey, Abhinav and Kadian, Abhishek and Al-Dahle, Ahmad and Letman, Aiesha and Mathur, Akhil and Schelten, Alan and Yang, Amy and Fan, Angela and others},
  journal={arXiv preprint arXiv:2407.21783},
  year={2024}
}

@article{wang2024preference,
  title={Preference Optimization with Multi-Sample Comparisons},
  author={Wang, Chaoqi and Zhao, Zhuokai and Zhu, Chen and Sankararaman, Karthik Abinav and Valko, Michal and Cao, Xuefei and Chen, Zhaorun and Khabsa, Madian and Chen, Yuxin and Ma, Hao and others},
  journal={arXiv preprint arXiv:2410.12138},
  year={2024}
}

@article{chen2021decision,
  title={Decision transformer: Reinforcement learning via sequence modeling},
  author={Chen, Lili and Lu, Kevin and Rajeswaran, Aravind and Lee, Kimin and Grover, Aditya and Laskin, Misha and Abbeel, Pieter and Srinivas, Aravind and Mordatch, Igor},
  journal={Advances in neural information processing systems},
  volume={34},
  pages={15084--15097},
  year={2021}
}

@article{achiam2023gpt,
  title={Gpt-4 technical report},
  author={Achiam, Josh and Adler, Steven and Agarwal, Sandhini and Ahmad, Lama and Akkaya, Ilge and Aleman, Florencia Leoni and Almeida, Diogo and Altenschmidt, Janko and Altman, Sam and Anadkat, Shyamal and others},
  journal={arXiv preprint arXiv:2303.08774},
  year={2023}
}

@inproceedings{
li2024hamster,
title={{HAMSTER}: Hierarchical Action Models for Open-World Robot Manipulation},
author={Yi Li and Yuquan Deng and Jesse Zhang and Joel Jang and Marius Memmel and Caelan Reed Garrett and Fabio Ramos and Dieter Fox and Anqi Li and Abhishek Gupta and Ankit Goyal},
booktitle={1st Workshop on X-Embodiment Robot Learning},
year={2024},
url={https://openreview.net/forum?id=yF3UekSJus}
}

@INPROCEEDINGS{12075,
  author={Kyriakopoulos, K.J. and Saridis, G.N.},
  booktitle={Proceedings. 1988 IEEE International Conference on Robotics and Automation}, 
  title={Minimum jerk path generation}, 
  year={1988},
  volume={},
  number={},
  pages={364-369 vol.1},
  doi={10.1109/ROBOT.1988.12075}}

@article{flash1985coordination,
  title={The coordination of arm movements: an experimentally confirmed mathematical model},
  author={Flash, Tamar and Hogan, Neville},
  journal={Journal of neuroscience},
  volume={5},
  number={7},
  pages={1688--1703},
  year={1985},
  publisher={Society for Neuroscience}
}

@inproceedings{mellinger2011minimum,
  title={Minimum snap trajectory generation and control for quadrotors},
  author={Mellinger, Daniel and Kumar, Vijay},
  booktitle={2011 IEEE international conference on robotics and automation},
  pages={2520--2525},
  year={2011},
  organization={IEEE}
}

@misc{liu2023liberobenchmarkingknowledgetransfer,
      title={LIBERO: Benchmarking Knowledge Transfer for Lifelong Robot Learning}, 
      author={Bo Liu and Yifeng Zhu and Chongkai Gao and Yihao Feng and Qiang Liu and Yuke Zhu and Peter Stone},
      year={2023},
      eprint={2306.03310},
      archivePrefix={arXiv},
      primaryClass={cs.AI},
      url={https://arxiv.org/abs/2306.03310}, 
}

@misc{fei2025liberoplusindepthrobustnessanalysis,
      title={LIBERO-Plus: In-depth Robustness Analysis of Vision-Language-Action Models}, 
      author={Senyu Fei and Siyin Wang and Junhao Shi and Zihao Dai and Jikun Cai and Pengfang Qian and Li Ji and Xinzhe He and Shiduo Zhang and Zhaoye Fei and Jinlan Fu and Jingjing Gong and Xipeng Qiu},
      year={2025},
      eprint={2510.13626},
      archivePrefix={arXiv},
      primaryClass={cs.RO},
      url={https://arxiv.org/abs/2510.13626}, 
}

\end{document}